\documentclass[letterpaper, 10 pt, conference]{ieeeconf}  

\IEEEoverridecommandlockouts                              

\usepackage{graphics} 
\usepackage{epsfig} 
\usepackage{mathptmx} 
\usepackage{times} 
\usepackage{amsmath} 
\usepackage{amssymb}  
\usepackage{color}
 \usepackage{cite}
 \usepackage{multirow,multicol}
 \usepackage{tabularx}
 \usepackage{siunitx}
 \usepackage{chemformula}

\usepackage[capitalize]{cleveref}

\title{\LARGE \bf
Martian Exploration of Lava Tubes (MELT) with ReachBot: Scientific Investigation and Concept of Operations
}

\author{Julia Di$^{1}$, Sara Cuevas-Qui\~{n}ones$^{2,3}$, Stephanie Newdick$^{4}$, \\ Tony G. Chen$^{1}$, Marco Pavone$^{4}$, Mathieu G. A. Lap\^{o}tre$^{3}$, Mark Cutkosky$^{1}$
\thanks{$^{1}$Department of Mechanical Engineering, Stanford University,
        Stanford, CA, 94035, USA
        {\tt\small juliadi@stanford.edu, agchen@stanford.edu, cutkosky@stanford.edu}}%
\thanks{$^{2}$Department of Planetary Sciences, Purdue University,
        West Lafayette, IN 47907, USA
        {\tt\small cuevas26@purdue.edu}}%
\thanks{$^{3}$Department of Earth and Planetary Sciences, Stanford University,
        Stanford, CA, 94035, USA
        {\tt\small mlapotre@stanford.edu}}
\thanks{$^{4}$Department of Aerospace Engineering, Stanford University,
        Stanford, CA, 94035, USA
        {\tt\small snewdick@stanford.edu, pavone@stanford.edu}}
}

\begin{document}

\maketitle
\thispagestyle{empty}
\pagestyle{empty}

\begin{abstract}
As natural access points to the subsurface, lava tubes and other caves have become premier targets of planetary missions for astrobiological analyses. Few existing robotic paradigms, however, are able to explore such challenging environments. ReachBot is a robot that enables navigation in planetary caves by using extendable and retractable limbs to locomote. This paper outlines the potential science return and mission operations for a notional mission that deploys ReachBot to a martian lava tube. In this work, the motivating science goals and science traceability matrix are provided to guide payload selection. A Concept of Operations (ConOps) is also developed for ReachBot, providing a framework for deployment and activities on Mars, analyzing mission risks, and developing mitigation strategies.

\end{abstract}



\section{INTRODUCTION}

Cave-like environments may hold the key to our search for life beyond Earth. As relatively stable micro-environments shielded from surface radiation, subsurface cavities could also act as a potential shelter for human habitation. More than a thousand cave-like features have been reported on Mars, including many thought to have formed from volcanic processes, such as lava tubes \cite{cushing2015atypical}. Importantly, a deep-drilling payload is not necessary to access some of these lava tubes; instead, skylights exist where portions of the tube ceiling did not properly form or collapsed. Due to the high scientific potential of accessible subsurface spaces, several missions and architectures have been proposed for their exploration, including the New Frontiers concept MACIE \cite{Phillips2021Mars}. 

\begin{figure}
    \centering
    \includegraphics[width=0.99\columnwidth]{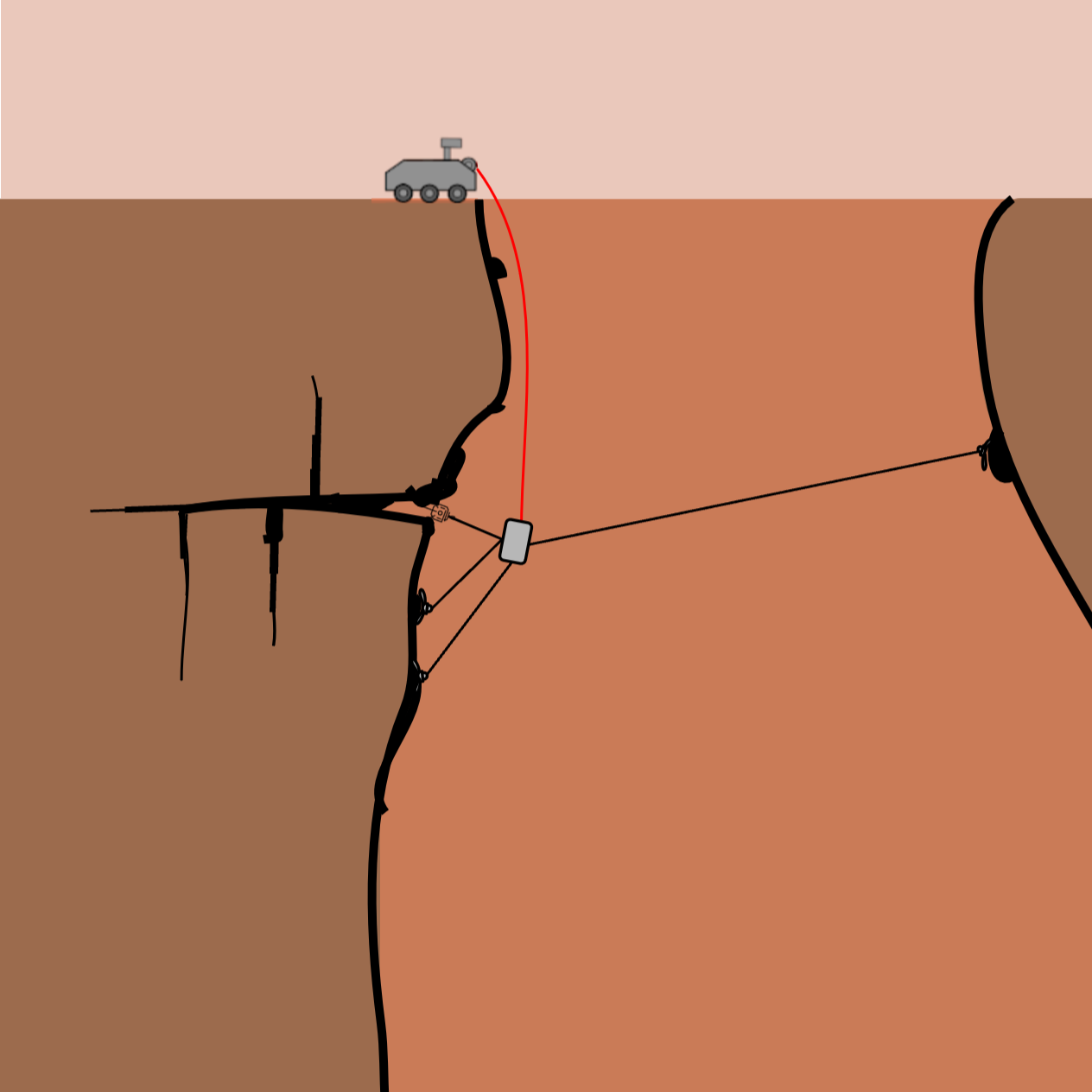}
    \caption{A depiction of a tethered ReachBot entering a lava tube via a skylight. ReachBot uses deployable booms as prismatic limbs, allowing the negotiation of sparse anchor points along the cliff and access to scientifically interesting areas such as stratigraphy along underhangs or fractures that would be inaccessible with other robot morphologies.} 
    \label{fig:reachbot}
    \vspace{-7mm}
\end{figure}

Robotic technologies that could access such environments include rock climbing robots (e.g., LEMUR \cite{parness2017lemur}), two-wheeled axial rappelling robots (e.g., MoonDiver \cite{nesnas2023moondiver} and DuAxel \cite{nesnas2012axel}), and drones or tumbling robots (e.g., Ingenuity \cite{balaram2021ingenuity} and Hedgehog \cite{ReidRovedaEtAl2014}). These architectures trade off between mobility and manipulation; typically a small and highly mobile system lacks 
the ability to exert substantial forces on the environment
while a larger system, suited for forceful interactions, may be prohibitively heavy for transport to remote sites and may have difficulty negotiating tight spaces.

Previous publications have introduced ReachBot, a robot that uses unrolling extendable space booms as limbs, each mounted to a pivoting shoulder. The extending limbs provide ReachBot with a couple of advantages for operating in spaces like caves and crevasses. First, they give ReachBot a very long reach for its body size, allowing it to use widely separated grasping sites (\cref{fig:reachbot}). 
Second, although the limbs are relatively weak in bending and compression, they can exert large tensile forces. Thus, a ReachBot with a small central body interacts forcefully with its environment.
%
%
These capabilities allow it to navigate to and perform operations at high-value science targets such as exposed stratigraphy and solidified lava flows that may otherwise be inaccessible. 

Parts of ReachBot's motion planning and control, perception, and hardware designs have been developed \cite{SchneiderBylardEtAl2022, NewdickOngoleEtAl2023,NewdickChenEtAl2023,DiEtAl2024,roberts2023skeleton, ChenMillerEtAl2022}, culminating in a recent system demonstration in a terrestrial lava tube of the Mojave Desert, California \cite{ChenNewdickEtAl2023}. Preliminary science mission concepts were also presented at the American Geophysical Union Fall Meeting in 2023 \cite{CuevasAGU}. This paper draws upon the existing body of work to (1) outline the potential science return of Martian Exploration of Lava Tubes (MELT), a  notional scientific mission with ReachBot, and (2) develop ReachBot's concept of operations (ConOps). 


\begin{table*}
    \centering
\caption{Science Traceability Matrix} 
\label{tab:sciencetrace}
    \begin{tabularx}{\linewidth}{ | c | c | c | c | } \hline
         \textbf{Goal} & \textbf{Science Objective} & \textbf{Observations and Measurements} & \textbf{Instruments} \\
\hline 
        \multirow{3}{3cm}{1. Assess the past and present habitability of a martian lava tube in situ.}
            & \multicolumn{1}{X|}{1A. Characterize the geological environment within the cave.} 
            & \multicolumn{1}{X|}{Images, chemical composition, and mineralogy within cave.} 
            & \multicolumn{1}{p{3cm}|}{Laser-Induced Breakdown Spectroscopy (LIBS), Raman spectroscopy, cameras, LiDAR}
            \\\cline{2-4}
        
            & \multicolumn{1}{X|}{1B. Determine if liquid water was ever present in the lava tube.} 
            & \multicolumn{1}{X|}{Distribution of water in subsurface and flow features.} 
            & \multicolumn{1}{p{3cm}|}{Raman / Infrared Reflectance Spectroscopy}
            \\\cline{2-4}
            
            & \multicolumn{1}{X|}{1C. Determine if there is an energy source present.} 
            & \multicolumn{1}{X|}{Distribution of nutrients. Mineralogy and composition, presence of CHNOPS.} 
            & \multicolumn{1}{p{3cm}|}{LIBS, Raman spectroscopy}
            \\\cline{2-4}
\hline 
        \multirow{3}{3cm}{2. Search for biosignatures of ancient or extant life in martian lava tubes.} 
            & \multicolumn{1}{X|}{2A. Determine whether there is evidence of bio alteration.} 
            & \multicolumn{1}{X|}{Determine presence of organic molecules.} 
            & \multicolumn{1}{p{3cm}|}{Time resolved florescence (TRF) spectroscopy}
            \\\cline{2-4} 
         
            & \multicolumn{1}{X|}{2B. Determine if there is morphological evidence of life.} 
            & \multicolumn{1}{X|}{Alteration minerals within cave, mineralogy.}  
            & \multicolumn{1}{p{3cm}|}{Micro-imager}
            \\\cline{2-4} 
         
            & \multicolumn{1}{X|}{2C. Determine if the lava tube contains any other biosignatures.} 
            & \multicolumn{1}{X|}{Sedimentary structures and textures, size and shape of potential biominerals, size and shape of potential cell-like structures or cell-like assemblages. Pores and fractures.} 
            & \multicolumn{1}{p{3cm}|}{TRF spectroscopy}
            \\\cline{2-4}
\hline
        \multirow{2}{3cm}{3. Characterize the environment within the cave to support future human exploration.}
            & \multicolumn{1}{X|}{3A. Determine if radiation levels in the lava tube are safe for microorganisms and human health.} 
            & \multicolumn{1}{X|}{Radiation levels in the lava tube.} 
            & \multicolumn{1}{p{3cm}|}{Radiation Assessment Detector (RAD)}
            \\\cline{2-4}

            & \multicolumn{1}{X|}{3B. Assess presence and usability of ice in the lava tube.} 
            & \multicolumn{1}{X|}{Mineralogy and location in the lava tube.} 
            & \multicolumn{1}{p{3cm}|}{Spectroscopy and camera}
            \\\cline{2-4}
\hline
    \end{tabularx} 
\end{table*}

\begin{table*}
    \centering
\caption{Instrument Mass and Power Budget}
\label{tab:payloadbudget}
    \begin{tabularx}{\textwidth}{|c|X|c|c|} \hline 
         \textbf{Instrument Suite} &  \textbf{Instrument(s)} & \textbf{Mass} (\SI{}{\kilo\gram})&  \textbf{Power} (Watts) \\ \hline 
         \multirow{5}{*}{SuperCam~\cite{wiens2021supercam}} & Optical Camera     & \multirow{5}{*}{10.6} & \multirow{5}{*}{17.9} \\
            & Laser-Induced Breakdown Spectroscopy (LIBS)  &  &  \\
            & Raman Spectroscopy (532 nm and can investigate targets up to \SI{12}{\meter} from the instrument)  &  &  \\
            & Time-Resolved Fluorescence (TRF) Spectroscopy  &  &  \\
            & Visible and Infrared Reflectance Spectroscopy  &  &  \\ \hline                        
         --- & Radiation Assessment Detector (RAD)&  1.5&  4.2\\ \hline 

         --- & High-intensity light source for spectroscopy & 0.2 & 40 (peak) \\ \hline
    \end{tabularx}
    
\end{table*}


\section{MELT Science Investigation}
\label{sec:science}

\subsection{Why Lava Tubes?}


Lava tubes are subsurface lava channels within which lava can maintain a viscosity low enough to flow significant distances\cite{cushing2012candidate,sakimoto1997eruption}. When the eruption ceases, the lava tubes may then empty, leaving behind intact and structurally stable tunnels. Surface entrances, known as skylights, may exist where portions of lava tube ceilings either did not fully form or have collapsed. Commonly observed in terrestrial basaltic settings, it is anticipated that lava tubes are also prevalent in volcanic landscapes on Mars \cite{sakimoto1997eruption,sauro2020lava,keszthelyi1995preliminary}.

The exploration of lava tubes on Mars could reveal profound insights about the solar system. On Earth, the subsurface harbors abundant microbial communities \cite{onstott2019paleo}, making these underground micro-environments promising locations for discovering evidence of past or present microbial life on Mars~\cite{leveille2010lava, Phillips2021Mars}. Furthermore, the subsurface could represent potential repositories for stable or metastable water-ice deposits; in particular, the highly insulated environment of lava tubes creates isolated microclimates by trapping and retaining cold air, which may lead to condensation for those that extend downward \cite{ingham2004climate}. Lava tube exploration also provides direct insight into lava flow dynamics on Mars and better constrains models for speleogenesis~\cite{wynne2022fundamental}. Finally, for human exploration, martian lava tubes emerge as a cost-effective in-situ choice for shielding humans from hazardous surface conditions.  

\subsection{MELT Science Goals and Objectives}

 
The high-level science goals of MELT are to: 
\begin{enumerate}
\item Assess the past and present habitability of a martian lava tube in situ.
\item Search for biosignatures of ancient or extant life in a martian lava tube. 
\item Characterize the environment within the cave to support future human exploration.
\end{enumerate} 

A successful mission with ReachBot is anticipated to address these goals. Understanding the present and past habitability of a lava tube (primary goal) is critical to providing context for any observed life-related processes or evidence (secondary goal). Previous studies have found geomorphic, mineralogical, and geochemical evidence that Mars was habitable in the Late Noachian-Early Hesperian~\cite{crater2014habitable,bosak2021searching,ehlmann2014mineralogy,hurowitz2017redox}. Furthermore, evidence of organics~\cite{scheller2022aqueous,house2022depleted}, and seasonal methane~\cite{webster2015mars} could also be signatures of extant life. In the case where no evidence of ancient or extant or life is detected, the habitability investigation also provides context for why evidence is absent. 

\subsection{Site Selection}
The Tharsis Montes region, specifically Arsia Mons, is thought to host an abundance of lava tubes, with hundreds of exposed skylights \cite{cushing2007themis}. Selecting Arsia Mons as a target would allow a reconnaissance mission to more efficiently evaluate lava tube candidates and focus on a specific target for a ReachBot mission. To maximize the likelihood of successfully accessing the subsurface, the mission focuses on the Tharsis region, which is clearly volcanic, of Amazonian age (young), and has clear skylights with little evidence for collapse~\cite{cushing2012candidate}. Older lava tubes are more likely to have hosted life, but are also more likely to have collapsed and be inaccessible for robots.

\subsection{Science Payload}

In the science traceability matrix, Table~\ref{tab:sciencetrace}, we map each science goal to required measurements and instruments. We also provide Table~\ref{tab:payloadbudget} for the payload mass and power budget. This payload mass is aligned with budgets developed in previous papers based on the load capacity of space booms \cite{NewdickChenEtAl2023, DiEtAl2024}. Below we discuss our science objectives and flight-heritage instruments that may satisfy the objectives.


\begin{itemize}
    \item \textbf{1A. Characterize the geological environment within the cave.} Imaging of cave walls, fractures, and other features such as lavacicles will provide insights into the formation and evolution of the lava tube, and context for assessing past habitability. Cameras and LiDAR may be used. Furthermore, it is likely that micro-organisms in a cave would have consisted of chemolithotrophs, such that compositional information (chemical and mineralogical, as measured with laser-induced breakdown spectroscopy (LIBS) and Raman spectroscopy) of the geological materials within the cave would shed light onto potential metabolic pathways, in addition to providing invaluable constraints on Mars' internal evolution.
    
    \item \textbf{1B. Determine if liquid water was ever present in the lava tube.} Liquid water is thought to be key to habitability, and could also be a resource for future human exploration. By providing relatively stable microclimates, caves may have hosted moisture at some point in Mars' history \cite{cushing2007themis}. Furthermore, hundreds of skylights in the Tharsis region have been remotely mapped and could hold metastable water or ice to this day \cite{cushing2015atypical, williams2010ice, Phillips2021Mars}. Whether in the form of liquid water, ice, or hydrated minerals,  techniques such as Raman or infrared reflectance spectroscopy (both parts of the SuperCam~\cite{wiens2021supercam} onboard the Perserverance rover) could be used to map the distribution of \ch{H2O} in the subsurface. 

    \item \textbf{1C. Determine if there is an energy source present.}
    The elements carbon, hydrogen, nitrogen, oxygen, phosphorus, and sulfur (CHNOPS) make up the majority of biomolecules on Earth. Elemental composition and mineralogical techniques (e.g., LIBS and Raman spectroscopy) help constrain the distribution and abundances of such elements within rocks and, possibly, ice.
    
\end{itemize}

We also seek to determine whether lava tubes were ever inhabited, leading to our secondary goal objectives: 

\begin{itemize}
    \item \textbf{2A. Identify traces of bio-alteration.}
    Bio-alteration of minerals (such as microbially mediated redox reactions) can lead to the formation of mineral deposits with distinct chemistry and morphologies\cite{posth2014biogenic,Phillips2021Mars}. Mineralogical techniques such as Raman and a micro-imager may be used to identify such deposits.

    \item \textbf{2B. Identify any morphological evidence for life.}
    A micro-imager can identify structures such as secondary porosity, fractures, and potential microfossils.

    \item \textbf{2C. Identify molecular biosignatures.}
    Time fluorescent spectroscopy, a part of SuperCam,  can detect organic molecules of biomolecular origin.
\end{itemize}

Finally, ReachBot would characterize the environment within the cave to support future crewed missions to Mars through the following objectives: 

\begin{itemize}
    \item \textbf{3A. Determine if radiation levels in the lava tube are safe for microorganisms and human health.} Although some specialized extremophiles can survive high exposures to radiation, many others (including humans) cannot. A radiation detector could quantify the level of radiation to provide clues to past habitability and assess the suitability of lava tubes as in-situ habitats.
    
    

    \item \textbf{3B. Assess presence and usability of ice in the lava tube.} Ice may be preserved for long timescales in martian cave environments, allowing possible extraction for crewed missions as drink or fuel. Spectroscopy techniques could confirm presence and abundance.
\end{itemize}

\section{MELT Mission Architecture}
Before ReachBot's mission, preliminary exploration of multiple Subsurface Access Points (SAPs) should confirm the presence of a lava tube and assess its physical conditions. Doing so will also help adjust ReachBot’s mobility capabilities to the specifics of the environment (e.g., robot scale and gripper technology). A Mars helicopter like Ingenuity is light enough, at only \SI{1.8}{\kilo\gram} \cite{balaram2021ingenuity}, to arrive with an earlier mission and explore several SAPs within \SI{100}{\kilo\meter} of each other. Candidate sites would be selected using the High-Resolution Imaging Experiment (HiRISE) camera onboard the Mars Reconnaissance Orbiter \cite{mcewen2007mars}. This reconnaissance helicopter, equipped with a camera, a LiDAR, and a visible-shortwave infrared (VSWIR) spectrometer, allows for initial evaluations of rock properties, texture, and composition. At least three potential SAPs should be mapped. The final lava tube for ReachBot ideally has a small entrance and an extended cavity that allows for the preservation of any present biosignatures as well as a large selection of explorable terrain. 

The configuration of ReachBot itself has been detailed in previous trade studies \cite{NewdickChenEtAl2023, DiEtAl2024}. 
Summarized here, ReachBot has up to eight booms attached to a central body containing engineering sensors (e.g., LiDAR and cameras) and scientific payload as described in Table~\ref{tab:payloadbudget}. ReachBot remains in development while its destination is chosen; the robot scale, including body diameter and maximum boom extension, can be adjusted based on the destination lava tube. 

ReachBot is deployed alongside a mothercraft rover that will remain anchored to the surface while ReachBot descends via tether. The rover and ReachBot will be deployed by an entry and descent landing platform, similar to the sky crane The rover is that it must have mobility capability to reach the entrance of the SAP, anchoring capability, tether deployment capability, and a power source such as a radioisotope thermoelectric generator.

\section{Concept of Operations}

\subsection{Modes of Operations}
\begin{figure}[h]
    \centering
    \includegraphics[width=1\columnwidth]{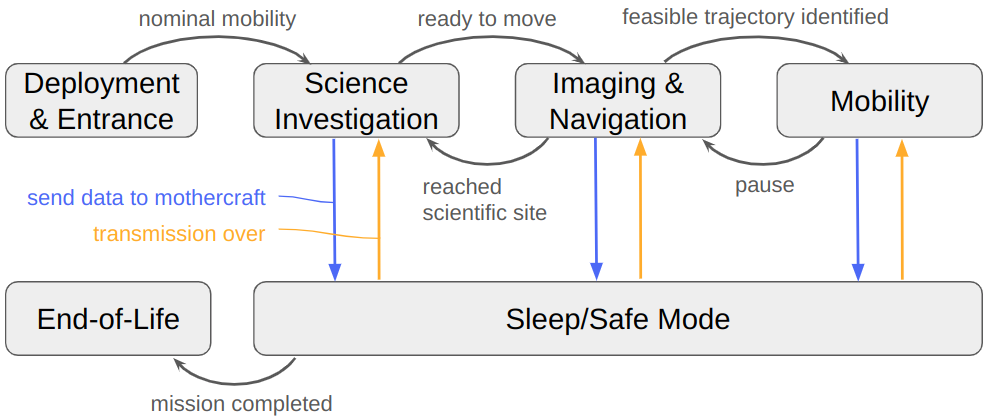}
    \vspace{-15pt}
    \caption{State diagram of ReachBot's six operational modes.} 
    \label{fig:modes_operation}
    \vspace{-15pt}
\end{figure}

The primary activities for MELT are Deployment \& Entrance, Science Investigation, Mobility, Imaging/Navigation, and Sleep/Safe Mode. Figure~\ref{fig:modes_operation} illustrates the state diagram that defines transitions between modes.

\begin{figure}
    \centering
    \includegraphics[width=1.0\columnwidth]{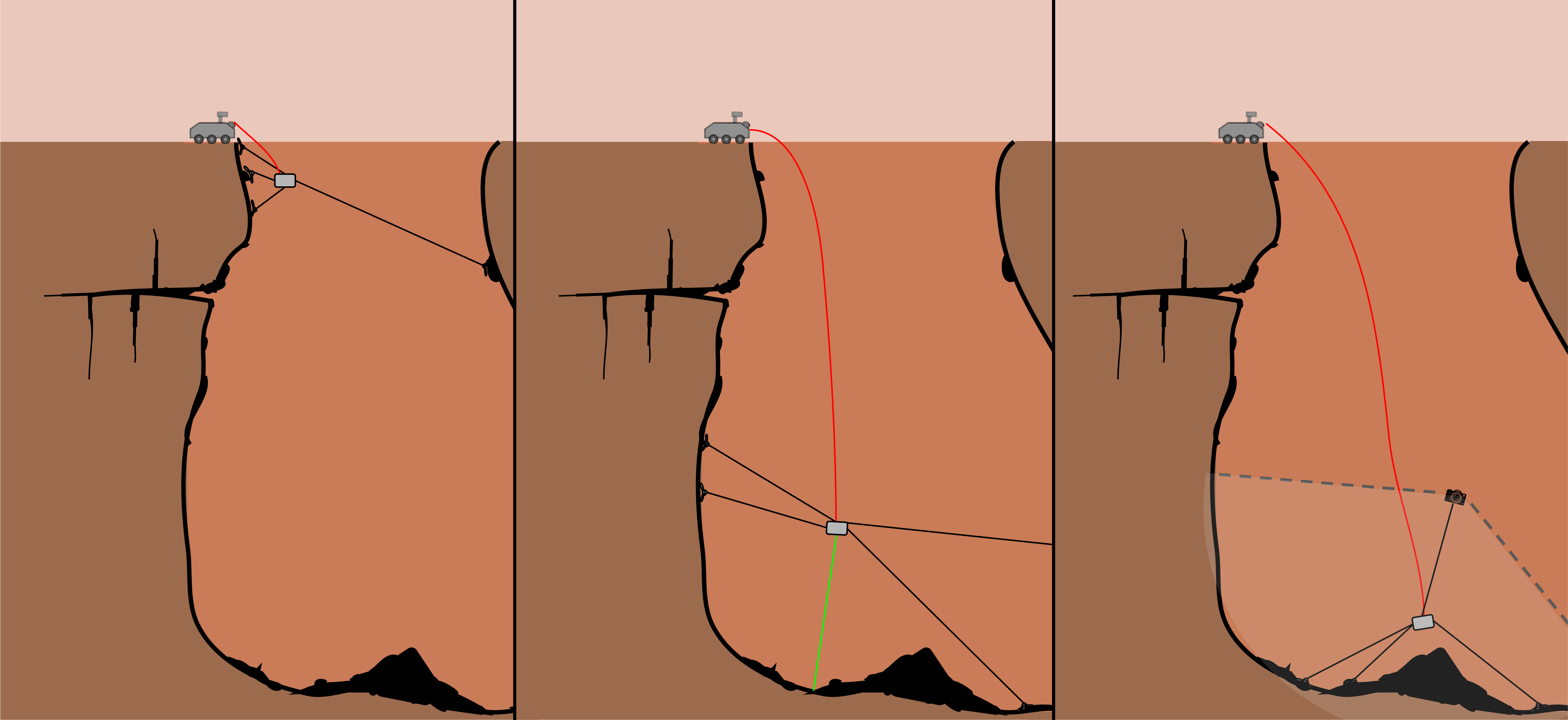}
    \caption{\textit{(From left to right)} ReachBot first deploys from the mothercraft, entering a lava tube via skylight by using a boom in tension on the opposing rock face; next, it reaches a scientifically-valuable area and collects spectroscopy measurements; it then uses remote imaging to inform navigation.} 
    \label{fig:entrance}
\end{figure}

Upon entering the cavity and conducting an initial, static imaging survey, ReachBot begins to move through the cavern using its extendable booms, interspersed with pauses for imaging and navigation. These intermissions are crucial for recalibrating the trajectory towards ReachBot's next science targets. Arrival at a primary science site triggers the next phase of in-depth scientific investigation, but scientific measurements are not limited to only these focused target sites. While primary scientific measurements and bulk data transmissions are reserved for the Science Investigation phase, additional images and measurements are continuously gathered throughout other phases to include waypoint finding. Each mode is described as follows:

\noindent\textbf{Deployment \& Entrance Phase: }
ReachBot is delivered to the surface via a primary spacecraft, such as a lander or rover, equipped with a precise landing system and/or transportation to the SAP. Upon positioning at the entrance, the mothercraft anchors to the surface and ReachBot undergoes self-checks and system diagnostics. To begin exploration, ReachBot first deploys from the mothercraft while remaining attached via tether. The tether should be actively controlled by ReachBot, for example via a caster arm assembly similar to that of the Axel rover~\cite{nesnas2012axel}. ReachBot is deployed from its stowed configuration on the underside, side, or top of this mothercraft. During descent into the lava tube, an initial imaging survey is conducted, including wall imaging with the SuperCam camera and engineering cameras, and secondary, continuous measurements of radiation levels with RAD with depth.

ReachBot's initial movements fall into one of two locomotion modes. The first mode is ReachBot's nominal locomotion, where the booms extend, grasp onto protruding rocks, and operate as tensile members. This mode is favorable, as it allows ReachBot to assume nominal locomotion immediately. However, it also requires that suitable grasp points are accessible from ReachBot's stowed position, e.g. on an opposing rock face as in Fig.~\ref{fig:entrance}.

If direct line-of-sight grasp points are not available, ReachBot might temporarily use its booms as legs. At short lengths, ReachBot's booms can support compressive loads, allowing operation as a legged robot. This capability allows ReachBot to deploy from the mothercraft similar to Ingenuity unfurling from the underside of Perseverance~\cite{balaram2021ingenuity}. In this configuration, ReachBot descends over the edge of the lava tube entrance by rapelling: the tether acts in tension, and at least some of ReachBot's booms act as ``legs" in compression holding the robot body away from the rock face. ReachBot can descend in this manner indefinitely, or until available grasp points are sufficient for a fully tensile configuration.

\noindent\textbf{Science Investigation: }
ReachBot conducts science investigations while stationary over areas of high scientific potential. 
As detailed in Sec.~\ref{sec:science}, ReachBot's payload includes SuperCam and RAD. In turn, SuperCam hosts a LIBS spectrometer (\SI{7}{\meter} range), a Raman spectrometer (\SI{12}{\meter} range), a time-resolved fluorescence (TRF) spectrometer, a visible and infrared spectrometer, and a micro-imager.

Once both scientific target and measurement are selected, the duration of the science phase is dictated by the necessary pre-cooling of SuperCam's Charge-Coupled Devices (CCDs) for 15 minutes before acquiring spectral data. Measurements themselves require only seconds, and analysis takes 4 minutes~\cite{wiens2021supercam}. The total energy reported for one observation point with SuperCam, including power-on and warm-up, is \SI{13.8}{\watt\hour}, with a total data volume of 26.5 Mbytes achieved with compression~\cite{wiens2021supercam}. While stationary, ReachBot transmits data at 10 Mbits/s with the High Speed Serial Link (HSSL) as currently done by the Perseverance rover~\cite{maurice2021supercam}. 



\noindent\textbf{Imaging/Navigation: }
Before and during ReachBot's mobility sequences, the robot uses its perception system to scan its environment. In particular, ReachBot uses a two-stage perceptual process: a cursory scan from the body provides rough indications of suitable grasp sites (e.g., convex features), then further surveying from an extended boom refines the grasp site evaluation. The specific instrumentation of this two-stage strategy is discussed in a prior trade study~\cite{DiEtAl2024}. 

ReachBot's mobility mode includes natural pauses between body movement and end-effector movement. These pauses serve multiple purposes: in addition to allowing time to identify grasp points for locomotion, they enable local data collection and transmission (both scientific analysis and localization). Planning could be executed autonomously or orchestrated based on instructions received from the ground operations team, depending on the mission's demands.

\noindent\textbf{Mobility: }
ReachBot's motion planning and locomotion have been presented in past work~\cite{NewdickOngoleEtAl2023,ChenNewdickEtAl2023}. 
To summarize, ReachBot moves by alternating two types of continuous movement: one in which all grippers remain attached to the environment, and one in which a gripper detaches and moves to a new grasp point. The motion planner relies on environmental scanning to quantify grasp sites and plan corresponding footsteps. 

By using extendable booms, ReachBot remains mobile even when the availability of grasp sites varies. However, the potential irregularity of grasp sites also prohibits ReachBot's motion planning approach from relying on a regular gait, introducing large variability in predicted traversal speeds. 


\noindent\textbf{Sleep/Safe Mode: }
ReachBot maintains communication with Earth via the mothercraft through relay satellites or direct links when possible.
Data, including high-resolution images, sensor readings, and scientific analyses, are transmitted to Earth for monitoring and analysis.

By allocating time in Sleep/Safe mode, at least one ground-in-the-loop cycle is allowed (relayed through the mothercraft) at the worst light time from Earth to Mars. This time is used as a buffer to plan mobility sequences autonomously or perform additional science sequences.

\noindent\textbf{End of Life: }
At end of life, ReachBot halts all operations and disables its articulation and active sensing, remaining at the bottom of the lava tube in a passively stable manner. Periodic change-detection measurements may be made once the primary investigation is conducted. The final data are transferred to the mothercraft.  

\section{Physical Environment and Environmental Impacts}




\subsection{Nominal Conditions}


Although no direct photographic evidence or measurements have been taken inside a martian lava tube to date, there have been external examinations of lava tubes on Arsia, Olympus, and Hadriaca volcanoes through the Context camera images (CTX), high-resolution stereo digital terrain models (DTMs), and the HiRISE camera \cite{sauro2020lava}. The physical size of ReachBot could be scaled up or down depending on the lava tube dimensions, but it is currently designed to accommodate an average depth of \SI{30}{\meter}. 
The interior walls might have a different surface roughness from terrestrial lava tubes due to different gravity and lava viscosity. These studies also suggest that there are probably boulders, rockfalls, and other debris that form mounds under skylights \cite{wilkens2009corona}.






\subsection{Off-Nominal Conditions}



Nominally, ReachBot is designed and tested for lava tubes similar to those found on Earth in terms of the range of surface roughness. There are, however, a few off-nominal conditions that may be specific to martian lava tubes. One is the possible presence of frost inside lava tubes in the Tharsis region \cite{williams2010ice}. Although this would have interesting astrobiological implications, complete ice coverage would be difficult for ReachBot's microspine grippers to navigate and may require a different gripper solution. Any ice would most probably be in the form of patches or wedges, which should not affect ReachBot's mobility because ReachBot can simply extend its booms to reach non-icy anchor points. 

Another off-nominal condition is an abundance of obsidian, leading to smooth, glassy surfaces. However, the obsidian forms from felsic (quartz and feldspar-rich) lavas and is thus not expected in the targeted basaltic lava flows. Furthermore, any glassy surfaces would most likely constitute localized outcrops that ReachBot can easily avoid. This is because glassy textures form when lava is cooled rapidly (typically when ``quenched" by water or ice). Because the proposed region of interest is on a volcanic edifice, deep aquifers are not expected. Although there is evidence for the past presence of glaciers in the area \cite{shean2007recent}, Arsia is a shield volcano, which is not a morphology characteristic of subglacial eruptions, thus any interactions between lava and glacial ice would be local.




\subsection{Planetary Protection}


The MELT mission accesses lava tubes, an area designated as a “special region,” and therefore is required to comply with Planetary Protection Category IVc as designated by the Committee on Space Research (COSPAR), World Space Council Planetary Protection Policy. Accordingly, the mission would ensure a total bioburden level of $< 1.5 \times 10^{-4}$ spores, as proposed for the similar MACIE mission \cite{phillips2020science}. 

\section{Risks and Potential Issues}
There are a few challenges that ReachBot must successfully address with fail-safe mechanisms and contingency plans. The biggest challenge is the unknown terrain in lava tubes, which is primarily addressed through ReachBot's system design. With its booms, ReachBot becomes a variable-sized robot that may squeeze through smaller shafts (as long as they are at least the width of the body), and traverse large voids and chambers. ReachBot's quasistatic mobility and passive stability also provides robustness to temporary mechanical failures or communication outages. As there is always a chance of a failed grasp, significant work has been conducted in robust motion planning \cite{NewdickOngoleEtAl2023}. Finally, in a truly adversarial terrain scenario, stationary science may still be conducted with ReachBot's payload through remote sensing. 

\section{Conclusions}

This paper outlines the science and operations for MELT, a notional mission to a martian lava tube using ReachBot technology. Specifically, the science investigation, science traceability matrix, and operational framework are developed. Not covered in this work are a detailed power budget, pre-ReachBot mission timeline, and the entry, descent, and landing (EDL) sequence; these are left as future directions for study. By using ReachBot's mobility and manipulation capabilities to enable the MELT mission concept, this work contributes significantly to understanding the geological history of Mars and the potential for past or present life in its subterranean environments.

\addtolength{\textheight}{-12cm}   




\section*{ACKNOWLEDGMENT}

Support for this work was provided by NASA under the Innovative Advanced Concepts program (NIAC) and by the Air Force under an STTR award with Altius Space Machines. S. Cuevas-Qui\~{n}ones was supported by the Stanford Sustainability Undergraduate Research in Geoscience and Engineering program. S. Newdick is supported by NASA Space Technology Graduate Research Opportunities (NSTGRO).


\bibliographystyle{IEEEtran}
\bibliography{root,ASL_papers}

\end{document}